\documentclass[10pt,twocolumn,letterpaper]{article}

\usepackage[pagenumbers]{cvpr} %

\usepackage{graphicx}
\usepackage{amsmath}
\usepackage{amssymb}
\usepackage{booktabs}
\usepackage{enumitem}
\usepackage{balance}
\usepackage{multirow}
\usepackage{xcolor}
\usepackage{soul}
\usepackage{overpic}
\usepackage{flushend}

\usepackage[pagebackref,breaklinks,colorlinks,bookmarks=false]{hyperref}

\usepackage{calc}
\newlength\myheight
\newlength\mydepth
\settototalheight\myheight{Xygp}
\newcommand*\inlinegraphics[1]{%
  \settototalheight\myheight{Xygp}%
  \settodepth\mydepth{Xygp}%
  \raisebox{-\mydepth}{\includegraphics[height=\myheight]{#1}}%
}

\usepackage[capitalize]{cleveref}
\crefname{section}{Sec.}{Secs.}
\Crefname{section}{Section}{Sections}
\Crefname{table}{Table}{Tables}
\crefname{table}{Tab.}{Tabs.}

\newcommand{\duet}{DUET}

\def\cvprPaperID{3269} %
\def\confName{CVPR}
\def\confYear{2022}

\begin{document}

\title{Think Global, Act Local: Dual-scale Graph Transformer for Vision-and-Language Navigation}

\author{Shizhe Chen$^{\dag}$, Pierre-Louis Guhur$^{\dag}$, Makarand Tapaswi$^{\ddag}$, Cordelia Schmid$^{\dag}$ and Ivan Laptev$^{\dag}$ \\
$^{\dag}$Inria, \'Ecole normale sup\'erieure, CNRS, PSL Research University $^{\ddag}$IIIT Hyderabad\\
{\tt\small \url{https://cshizhe.github.io/projects/vln_duet.html}}
}

\pagestyle{plain}

\twocolumn[{%
\renewcommand\twocolumn[1][]{#1}%
\vspace{-0.5cm}
\maketitle

\begin{center}
\includegraphics[width=\linewidth]{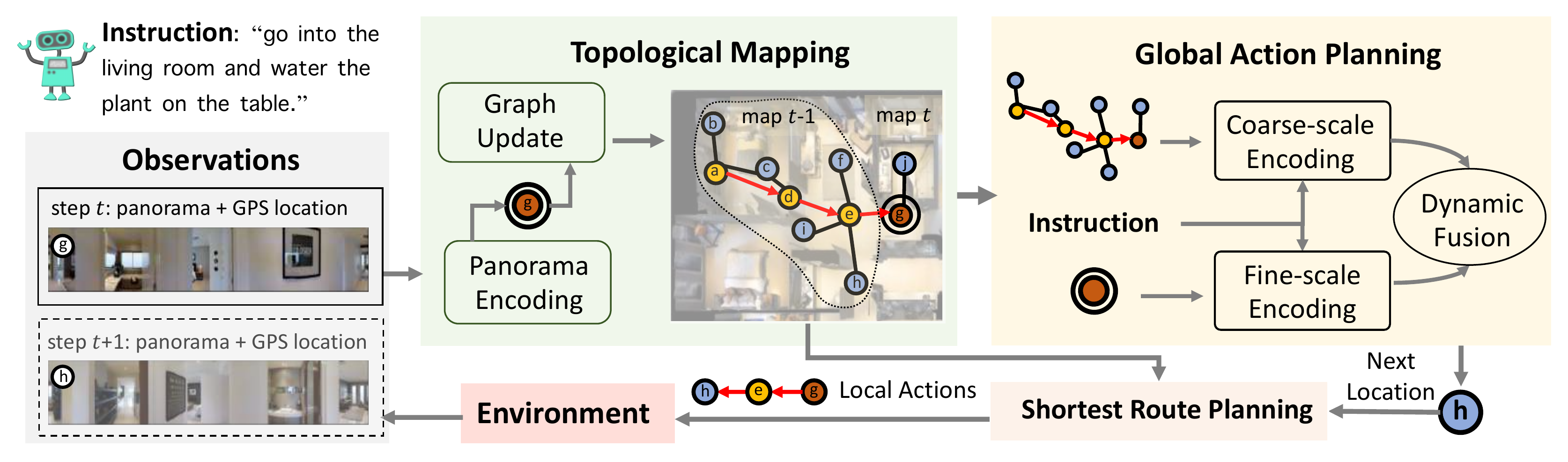}
\captionof{figure}{\small An agent is required to navigate in unseen environments to reach target locations according to language instructions. It only obtains local observations of the environment and is allowed to make local actions, \ie, moving to neighboring locations. In this work, we propose to build topological maps on-the-fly to enable long-term action planning. The map contains visited nodes~\includegraphics[width=1em]{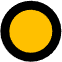} and navigable nodes~\includegraphics[width=1em]{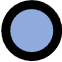} that can be reached from the previously visited nodes. 
Our method predicts global actions, \ie, all navigable nodes in the map, and trades off complexity by combining a coarse-scale graph encoding with a fine-scale encoding ~\includegraphics[width=1em]{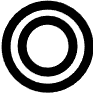} of observations at the current node~\includegraphics[width=1em]{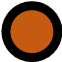}. 
}
\label{fig:teaser}
\end{center}
}]

\begin{abstract}
Following language instructions to navigate in unseen environments is a challenging problem for autonomous embodied agents.
The agent not only needs to ground languages in visual scenes, but also should explore the environment to reach its target.
In this work, we propose a {\bf du}al-scal{\bf e} graph {\bf t}ransformer (\duet) for joint long-term action planning and fine-grained cross-modal understanding. 
We build a topological map on-the-fly to enable efficient exploration in global action space.
To balance the complexity of large action space reasoning and fine-grained language grounding, we dynamically combine a fine-scale encoding over local observations and a coarse-scale encoding on a global map via graph transformers.
The proposed approach, \duet, significantly outperforms state-of-the-art methods on goal-oriented vision-and-language navigation (VLN) benchmarks REVERIE and SOON. It also improves the success rate on the fine-grained VLN benchmark R2R.
\end{abstract}
\vspace{-2em}

\section{Introduction}

Autonomous navigation is an essential ability for intelligent embodied agents.
Given the convenience of natural language for human-machine interaction, autonomous agents should also be able to understand and act according to human instructions. Towards this goal, Vision-and-Language Navigation (VLN)~\cite{anderson2018evaluation} is a challenging problem that has attracted a lot of recent research~\cite{anderson2018vision,chen2019touchdown,ku2020room,krantz2020beyond,shridhar2020alfred,qi2020reverie,zhu2021soon,irshad2021robovln}.
VLN requires an agent to follow language instructions and to navigate in unseen environments to reach a target location.
Initial approaches to VLN~\cite{anderson2018vision,chen2019touchdown,ku2020room} use fine-grained instructions providing step-by-step navigation guidance such as ``\emph{Walk out of the bedroom. Turn right and walk down the hallway. At the end of the hallway turn left. Walk in front of the couch and stop}''. 
This fine-grained VLN task enables grounding of detailed instructions but is less practical due to the need of step-by-step guidance. 
A more convenient interaction with agents can be achieved by goal-oriented instructions~\cite{qi2020reverie,zhu2021soon} such as ``\emph{Go into the living room and water the plant on the table}''.
This task, however, is more challenging as it requires both the grounding of rooms and objects as well as the efficient exploration of environments to reach the target.

In order to efficiently explore new areas, or
correct previous decisions,
an agent should keep track of already executed instructions and visited locations in its memory.
Many existing VLN approaches~\cite{anderson2018vision,fried2018speaker,ma2019self,wang2019reinforced,tan2019learning,hong2020recurrent} implement memory using recurrent architectures, \eg~LSTM, and condense navigation history in a fixed-size vector. Arguably, such an implicit memory mechanism can be inefficient to store and utilize previous experience with a rich space-time structure. 
A few recent approaches \cite{chen2021hamt, pashevich2021episodic} propose to explicitly store previous observations and actions, and to model long-range dependencies for action prediction via transformers \cite{vaswani2017attention}.
However, these models only allow for local actions, \ie, moving to neighboring locations. As a result, an agent has to run its navigation model $N$ times to backtrack $N$ steps, which increases instability and compute. 

\begin{figure} 
\centering
\begin{overpic}[width=1\linewidth]{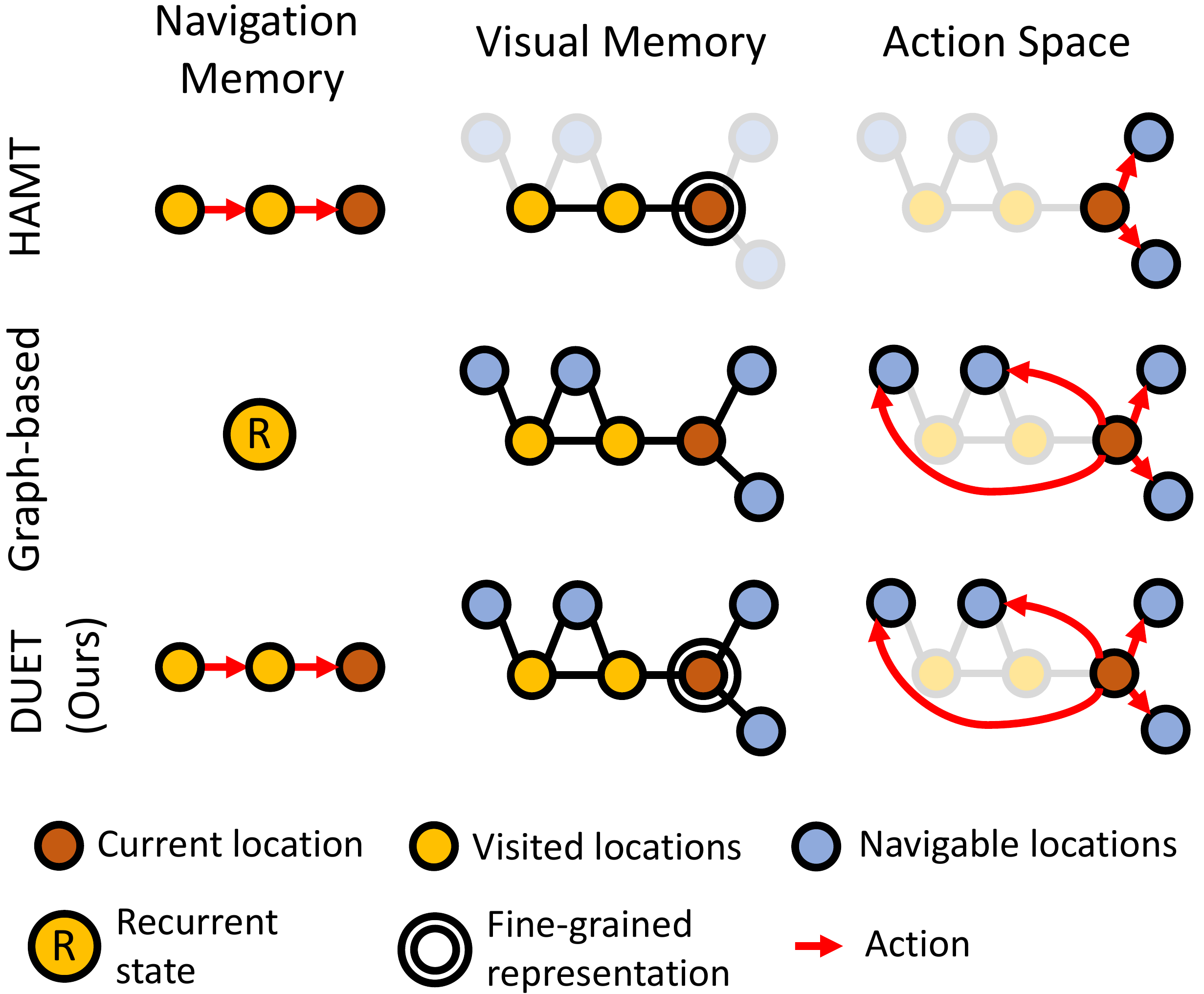}
\put(5.5,61){\rotatebox{90}{\cite{chen2021hamt}}}
\put(5.5,36){\rotatebox{90}{\cite{deng2020evolving,wang2021structured,zhu2021soon}}}
\end{overpic}
\caption{Method comparison. HAMT \cite{chen2021hamt} stores navigation and visual memories to capture long-range dependency in action prediction, but is limited to a local action space. Graph-based approaches \cite{deng2020evolving,wang2021structured,zhu2021soon} use topological maps to support a global action space, but suffer from a recurrent navigation memory and a coarse-scale visual representation. Our \duet~model overcomes previous limitations with a dual-scale encoding over the map.}
\label{fig:method_cmpr}
\vspace{-1em}
\end{figure}

A potential solution is to build a map \cite{chaplot2020neural} that explicitly keeps track of all visited and navigable locations observed so far. 
The map allows an agent to make efficient long-term navigation plans. For example, the agent is able to select a long-term goal from all navigable locations in the map, and then uses the map to calculate a shortest path to the goal. 
Topological maps have been explored by previous VLN works~\cite{deng2020evolving,wang2021structured,zhu2021soon}. These methods, however, still fall short in two aspects.
Firstly, they rely on recurrent architectures to track the navigation state as shown in the middle of Figure~\ref{fig:method_cmpr}, which can greatly hinder the long-term reasoning ability for exploration.
Secondly, each node in topological maps is typically represented by condensed visual features. Such coarse representations reduce complexity but may lack details to ground fine-grained object and scene descriptions in instructions.

Our approach addresses both of these shortcomings, the first one based on a transformer architecture and the second one with a dual-scale action planning approach. 
We propose a {\bf Du}al-scal{\bf e} graph {\bf T}ransformer (\duet) with topological maps. %
As illustrated in Figure~\ref{fig:teaser}, our model consists of two modules: topological mapping and global action planning.
In topological mapping, we construct a topological map over time by adding newly observed locations to the map and updating visual representations of nodes.
Then at each step, the global action planning module predicts a next location in the map or a stop action.
To balance fine-grained language grounding and reasoning over large graphs, we propose to dynamically fuse action predictions from dual scales: a fine-scale representation of the current location and a coarse-scale representation of the map.
In particular, we use transformers to capture cross-modal vision-and-language relations, and improve the map encoding by introducing the knowledge of graph topology into transformers. 
We pretrain the model with behavior cloning and auxiliary tasks, and propose a pseudo interactive demonstrator to further improve policy learning.
\duet~significantly outperforms state-of-the-art methods on goal-oriented VLN benchmarks REVERIE and SOON. It also improves success rate on fine-grained VLN benchmark R2R.
In summary, the contributions of our work are three-fold:
\parskip=0.1em
\begin{itemize}[itemsep=0.1em,parsep=0em,topsep=0em,partopsep=0em]
	\item We propose a dual-scale graph transformer (\duet) with topological maps for VLN. It combines coarse-scale map encoding and fine-scale encoding of the current location for efficient planning of global actions.
	\item We employ graph transformers to encode the topological map and to learn cross-modal relations with the instruction, so that action prediction can rely on a long-range navigation memory.
	\item \duet~achieves state of the art on goal-oriented VLN benchmarks, with more than 20\% improvement on success rate (SR) on the challenging REVERIE and SOON datasets. It also generalizes to fine-grained VLN task, \ie, increasing SR on R2R dataset by 4\%.
\end{itemize}

\section{Related work}
\label{sec:related_work}

\noindent\textbf{Vision-and-language navigation (VLN).}
Navigation tasks involving instruction following \cite{anderson2018vision, ku2020room, jain2019stay, chen2019touchdown,krantz2020beyond, irshad2021robovln,das2018embodied, yu2019multi,shridhar2020alfred} have become increasingly popular. 
Initial VLN methods mainly adopt recurrent neural networks with cross-modal attention~\cite{anderson2018vision, fried2018speaker, tan2019learning, ma2019regretful, vasudevan2021talk2nav}.
More recently, transformer-based architectures have been shown successful in VLN tasks~\cite{hao2020towards}, notably by leveraging pre-trained architectures.
For example, PRESS~\cite{li2019robust} adopts BERT~\cite{devlin2019bert} for instruction encoding. Different variants of ViLBERT are used in~\cite{majumdar2020improving, guhur2021airbert} to measure compatibility between instructions and visual paths, but cannot be used for sequential action prediction.
Recurrent VLN-BERT~\cite{hong2020recurrent} addresses the limitation by injecting a recurrent unit in transformer architectures for action prediction.
Instead of relying on one recurrent state, E.T.~\cite{pashevich2021episodic} and HAMT~\cite{chen2021hamt} directly use transformers to capture long-range dependency to all past observations and actions (see first row in Figure~\ref{fig:method_cmpr}).

\noindent\textbf{Maps for navigation.}
The work on visual navigation has a long tradition of using SLAM~\cite{thrun2002probabilistic} to construct metric maps \cite{thrun1998learning} of the environment, using non-parametric methods~\cite{huang2017visual}, neural networks~\cite{zhang2017neural, gupta2017cognitive}, or a mixture of both~\cite{chaplot2020learning}. Anderson~\etal~\cite{anderson2019chasing} employ such metric maps for VLN tasks.
However, it is challenging and requires accurate determination to construct metric map in real-time navigation.
Therefore, several works~\cite{savinov2018semi, fang2019scene} propose to represent the map as topological structures for pre-exploring environments~\cite{chen2021topological}, or for back-tracking to other locations, trading-off navigation accuracy with the path length~\cite{fried2018speaker, ma2019regretful}.
A few recent VLN works \cite{deng2020evolving,wang2021structured,zhu2021soon} used topological maps to support global action planning, but they suffer from using recurrent architectures for state tracking and also lack a fine-scale representation for language grounding as shown in Figure~\ref{fig:method_cmpr}.
We address the above limitations via a dual-scale graph transformer with topological maps.

\noindent\textbf{Training algorithms for sequential prediction.}
Behavior cloning is the most widely used training algorithm for sequential prediction. Nevertheless, it suffers from distribution shifts between training and testing.
To address the limitation, different training algorithms have been proposed such as scheduled sampling \cite{bengio2015scheduled}, DAgger \cite{ross2011reduction}, reinforcement learning (RL) \cite{sutton2018reinforcement}. 
Most VLN works \cite{tan2019learning,hong2020recurrent} combine behavior cloning and A3C RL \cite{mnih2016asynchronous}.
Wang \etal \cite{wang2020soft} propose to learn rewards via soft expert distillation.
Due to the difficulty of using RL in tasks with sparse rewards, we instead use an interactive demonstrator to mimic an expert and provide supervision in sequential training.

\section{Method}

\noindent\textbf{Problem formulation.}
In the standard VLN setup for discrete environments~\cite{anderson2018vision,qi2020reverie,zhu2021soon},
the environment is an undirected graph $\mathcal{G}=\{\mathcal{V}, \mathcal{E}\}$, where $\mathcal{V}=\{V_i\}_{i=1}^{K}$ denotes $K$ navigable nodes, and $\mathcal{E}$ denotes connectivity edges.
An agent is equipped with an RGB camera and a GPS sensor, and is initialized at a starting node in a previously unseen environment.
The goal of the agent is to interpret natural language instructions and to traverse the graph to the target location and find the object specified by the instruction.
$\mathcal{W}=\{w_i\}_{i=1}^{L}$ are word embeddings of the instruction with $L$ words.
At each time step $t$, the agent receives a panoramic view and position coordinates of its current node $V_t$.
The panorama is split into $n$ images $\mathcal{R}_t = \{r_i\}_{i=1}^{n}$, each represented by an image feature vector $r_i$ and a unique orientation.
To enable fine-grained visual perception, $m$ object features $\mathcal{O}_t=\{o_i\}_{i=1}^m$ are extracted in the panorama using annotated object bounding boxes or automatic object detectors~\cite{anderson2018bottom}.
In addition, the agent is aware of a few navigable views corresponding to its neighboring nodes $\mathcal{N}(V_t)$ as well as their coordinates.
The navigable views of $\mathcal{N}(V_t)$ are a subset of $\mathcal{R}_t$.
The possible local action space $\mathcal{A}_t$ at step $t$ contains navigating to $V_i \in \mathcal{N}(V_t)$ and stopping at $V_t$.
After the agent decides to stop at a location, it needs to predict the location of the target object in the panorama.

Exploration and language grounding are two essential abilities for VLN agents.
However, existing works either only allow for local actions $\mathcal{A}_t$ \cite{tan2019learning,hong2020recurrent,chen2021hamt} which hinders long-range action planning, or lack object representations $\mathcal{O}_t$ \cite{deng2020evolving,wang2021structured,zhu2021soon} which might be insufficient for fine-grained grounding. Our work addresses both issues with a dual-scale representation and global action planning.

\noindent\textbf{Overview.}
As illustrated in Figure~\ref{fig:teaser}, our model consists of two learnable modules, namely topological mapping and global action planning.
The topological mapping module gradually constructs a topological map over time. The global action planning module then performs dual-scale reasoning based on coarse-scale global observations and fine-scale local observations. 
In the following, we introduce topological mapping in Sec.~\ref{sec:method_mapping} and global action planning in Sec.~\ref{sec:method_global_planning}.
We end this section by presenting our approach to train our model and use it for inference in Sec.~\ref{sec:method_training}.

\subsection{Topological Mapping}
\label{sec:method_mapping}

\begin{figure}[t]
    \centering
    \includegraphics[width=1\linewidth]{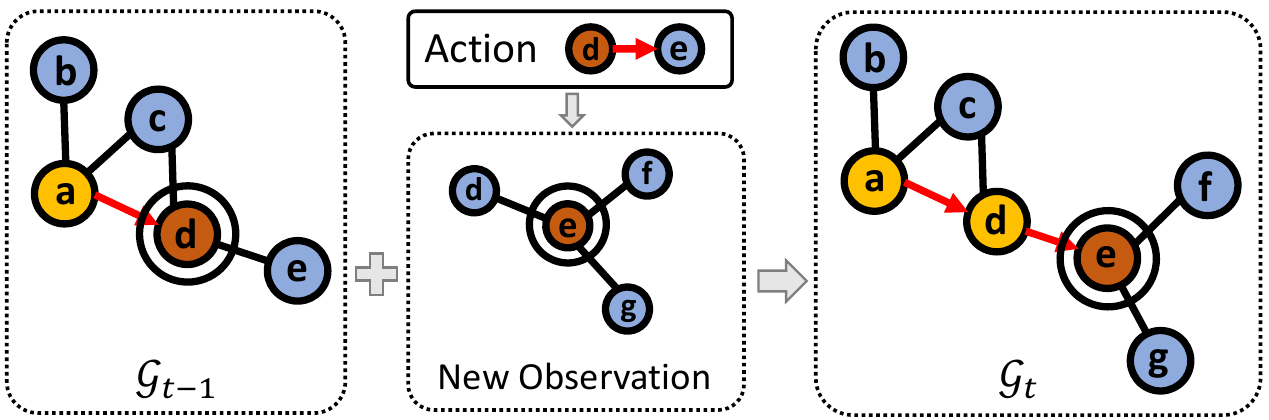}
    \caption{Illustration of graph updating at time step $t$. Given a new action $d \rightarrow e$, an agent receives new observations at node $e$. It then adds new nodes and updates node representations.}
    \label{fig:method_graph_update}
\end{figure}

The environment graph $\mathcal{G}$ is initially unknown to the agent, hence, our model gradually builds its own map using observations along the path.
Let $\mathcal{G}_t=\{\mathcal{V}_t, \mathcal{E}_t\}$ with $K_t$ nodes, $\mathcal{G}_t \subset \mathcal{G}$ be the map of the environment observed after $t$ navigation steps.
There are three types of nodes in $\mathcal{V}_t$ (see Figure~\ref{fig:teaser}): (i) visited nodes~\inlinegraphics{figures/visited_node}; (ii) navigable nodes~\inlinegraphics{figures/navigable_node}; and (iii) the current node~\inlinegraphics{figures/current_node}.
The agent has access to panoramic views for visited nodes and the current node. Navigable nodes are unexplored and are only partially observed from already visited locations, hence, they have different visual representations.
At each step $t$, we add the current node $V_t$ and its neighboring unvisited nodes $\mathcal{N}(V_t)$ to $\mathcal{V}_{t-1}$, and update $\mathcal{E}_{t-1}$ accordingly as illustrated in Figure~\ref{fig:method_graph_update}.
Given the new observation at $V_t$, we also update visual representations of the current node and navigable nodes as follows.

\noindent\textbf{Visual representations for nodes.}
At time step $t$, the agent receives image features $\mathcal{R}_t$ and object features $\mathcal{O}_t$ of node $V_t$.
We use a multi-layer transformer~\cite{vaswani2017attention} to model spatial relations among images and objects.
The core of the transformer is the self-attention block:
\begin{gather}
    [\mathcal{R}'_t, \mathcal{O}'_t] = \text{SelfAttn}\left([\mathcal{R}_t, \mathcal{O}_t] \right),\\
    \text{SelfAttn}(X) = \text{Softmax}\left(\frac{XW_q (XW_k)^{T}}{\sqrt{d}}\right) XW_v,
    \label{eqn:attn}
\end{gather}
where $W_* \in \mathbb{R}^{d \times d}$ are parameters and biases are omitted.
For ease of notation, we still use $\mathcal{R}_t, \mathcal{O}_t$ in the following instead of $\mathcal{R}'_t, \mathcal{O}'_t$ to denote the encoded embeddings.

Then we update visual representation of the current node~\inlinegraphics{figures/current_node} by average pooling of $\mathcal{R}_t$ and $\mathcal{O}_t$.
As the agent also partially observes $\mathcal{N}(V_t)$ at $V_t$, we accumulate visual representations of these navigable nodes~\inlinegraphics{figures/navigable_node} based on the corresponding view embedding in $\mathcal{R}_t$.
If a navigable node has been seen from multiple locations, we average all the partial view embeddings as its visual representation.
We use $v_i$ to denote the pooled visual representation for each node $V_i$.
Such a coarse-scale representation enables efficient reasoning over large graphs, but may not provide sufficient information for fine-grained language grounding especially for objects.
Therefore, we keep $\mathcal{R}_t, \mathcal{O}_t$ as a fine-grained visual representation~\inlinegraphics{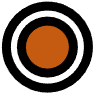} for the current node $V_t$ to support detailed reasoning at a fine-scale.

\begin{figure*}[t]
	\centering
	\includegraphics[width=1\linewidth]{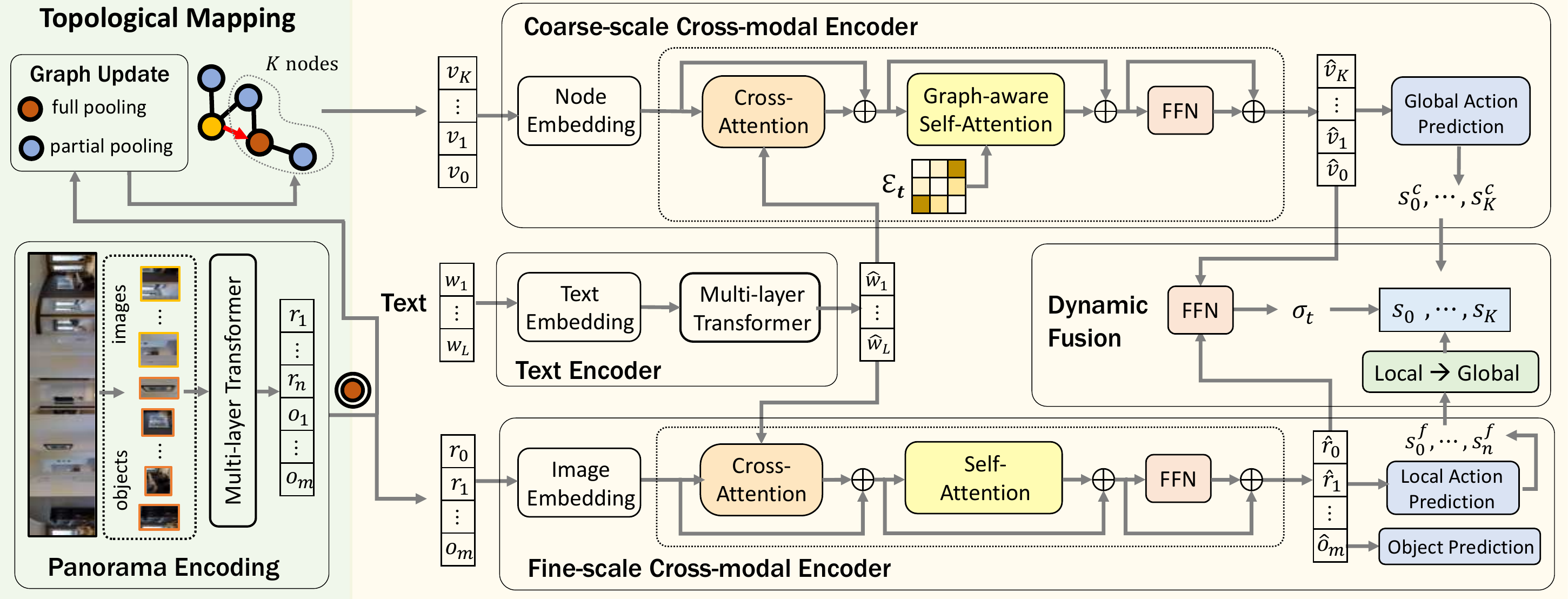}
	\caption{\duet~consists of topological mapping (left) and global action planning (right). The mapping module outputs a graph with $K$ node features $\{v_i\}_{i=1}^{K}$, and the current panorama encoding with image features $\{r_i\}_{i=1}^{n}$ and object features $\{o_i\}_{i=1}^{m}$. Node feature $v_0$ and image feature $r_0$ are used to indicate the `stop' action. The global action planning uses transformers for coarse- and fine-scale cross-modal encoding and fuses the two scales to obtain a global action score $s_i$ for each node.
	}
	\label{fig:method_global_action}
	\vspace{-0.8em}
\end{figure*}

\subsection{Global Action Planning}
\label{sec:method_global_planning}

Figure~\ref{fig:method_global_action} illustrates the global action planning module.
The coarse-scale encoder makes predictions over all previously visited nodes, but uses a coarse-scale visual representation. The fine-scale encoder instead predicts local actions given fine-grained visual representations of the current location.
The dynamic fusion of both encoders combines predictions of global and local actions. %

\vspace{-1.2em}
\subsubsection{Text Encoder} 
\label{sec:txt_encoding}
\vspace{-0.5em}
To each word embedding in $\mathcal{W}$ is added a positional embedding~\cite{devlin2019bert} corresponding to the position of the word in the sentence and a type embedding for text~\cite{tan2019lxmert}. 
All word tokens are then fed into a multi-layer transformer to obtain contextual word representations, denoted here as $\hat{\mathcal{W}} = \{\hat{w}_1, \cdots, \hat{w}_{L}\}$.

\vspace{-1.2em}
\subsubsection{Coarse-scale Cross-modal Encoder} 
\vspace{-0.5em}
\label{sec:coarse_scale_encoding}
The module takes the coarse-scale map $\mathcal{G}_t$ and encoded instruction $\hat{\mathcal{W}}$ to make navigation predictions over a global action space ($\cup_{i=1}^{t} \mathcal{A}_i$).

\noindent\textbf{Node embedding.}
To the node visual feature $v_i$ is added a location encoding and a navigation step encoding.
The location encoding embeds the location of a node in the map in an egocentric view, which is the orientation and distance relative to the current node.
The navigation step encoding embeds the latest visited time step for visited nodes and 0 for unexplored nodes. In this way, visited nodes are encoded with a different navigation history to improve alignment with the instruction.
We add a `stop' node $v_0$ in the graph to denote a stop action and connect it with all other nodes. 

\noindent\textbf{Graph-aware cross-modal encoding.}
The encoded node and word embeddings are fed into a multi-layer graph-aware cross-modal transformer.
Each transformer layer consists of a cross-attention layer \cite{tan2019lxmert} to model relations between nodes and instructions, and a graph-aware self-attention layer to encode environment layout.
The standard attention in Eq.~(\ref{eqn:attn}) only considers visual similarity among nodes, and thus it might overlook nearby nodes which are more relevant than distant nodes. To address the problem, we propose the graph-aware self-attention (GASA) which further takes into account the structure of the graph to compute attention as follows:
\begin{gather}
\label{eqn:graphaware_attn}
\small
    \text{GASA}(X) = \text{Softmax}\left(\frac{XW_q (XW_k)^{T}}{\sqrt{d}} + M\right) XW_v,\\
    M = E W_e + b_e,
\end{gather}
where $X$ denotes node representations, $E$ is the pair-wise distance matrix obtained from $\mathcal{E}_t$, and $W_e, b_e$ are two learnable parameters.
We stack $N$ layers in the encoder and denote the output embedding of node $V_i$ as $\hat{v}_i$.

\noindent\textbf{Global action prediction.}
We predict a navigation score for each node $V_i$ in $\mathcal{G}_t$ as below:
\begin{equation}
    \label{eqn:action_prediction}
    s^c_{i} = \text{FFN}(\hat{v}_i),
\end{equation}
where $\text{FFN}$ denotes a two-layer feed-forward network. 
To be noted, $s^c_0$ is the stop score. In most VLN tasks, it is not necessary for an agent to revisit a node, and thus we mask the score for visited nodes if not specially mentioned.

\vspace{-1.2em}
\subsubsection{Fine-scale Cross-modal Encoder} 
\label{sec:fine_scale_encoding}
\vspace{-0.5em}
This part attends to the current location $V_t$ in the map to enable fine-scale cross-modal reasoning.
The input is the instruction $\hat{\mathcal{W}}_t$ and fine-grained visual representations $\{\mathcal{R}_t, \mathcal{O}_t\}$ of the current node.
The module predicts navigation actions in a local action space ($\mathcal{A}_t$), and grounds the object at the final time step.

\noindent\textbf{Visual Embedding.}
We add two types of location embeddings to $\mathcal{R}_t, \mathcal{O}_t$.
The first type is the current location in the map relative to the start node. This embedding helps understand absolute locations in instruction such as ``\emph{go to the living room in first floor}''.
Then for $V_i \in \mathcal{N}(V_t)$, we add a second location embedding, the relative position of each neighboring node to the current node. It helps the encoder to realize egocentric directions such as ``\emph{turn right}''.
A special `stop' token $r_0$ is added for stop action.

\noindent\textbf{Fine-grained cross-modal reasoning.} 
We concatenate $[r_0; \mathcal{R}_t; \mathcal{O}_t]$ as visual tokens and exploit a standard multi-layer cross-modal transformer \cite{tan2019lxmert} to model vision and language relations.
The output embeddings of visual tokens are represented as $\hat{r}_0, \hat{\mathcal{R}}_t, \hat{\mathcal{O}}_t$ respectively.

\noindent\textbf{Local action prediction and object grounding.}
We predict a navigation score $s^f_{i}$ in local action space $\mathcal{A}_t$ similar to Eq.~(\ref{eqn:action_prediction}).
Moreover, as the goal-oriented VLN task requires object grounding, we further use a FFN to generate object scores based on $\hat{\mathcal{O}}_t$.

\vspace{-1.2em}
\subsubsection{Dynamic Fusion}
\vspace{-0.5em}
We propose to dynamically fuse coarse- and fine-scale action predictions for better global action prediction.
However, the fine-scale encoder predicts actions in a local action space which does not match with the coarse-scale encoder.
Therefore, we first convert local action scores $s^f_{i} \in \{\text{stop}, \mathcal{N}(V_t)\}$ into the global action space.
In order to navigate to other unexplored nodes that are not connected with the current node, the agent needs to backtrack through its neighboring visited nodes.
Therefore, we sum over scores of visited nodes in $\mathcal{N}(V_t)$ as an overall backtrack score $s_{\text{back}}$.
We keep the values for $s^f_{i} \in \{\text{stop}, \mathcal{N}(V_t)\}$ and 
use the constant $s_{\text{back}}$ for the others. 
Hence, the converted global action scores are:
\begin{equation}
\label{eq6}
s^{f'}_{i} =\left\{
\begin{aligned}
& s_{\text{back}},~\text{if}~V_i \in \mathcal{V}_t - \mathcal{N}(V_t), \\
& s^f_{i},~\text{otherwise}.
\end{aligned}
\right.
\end{equation}
At each step, we concatenate $\hat{v}_0$ from coarse-scale encoder and $\hat{r}_0$ from fine-scale encoder to predict a scalar for fusion:
\begin{equation}
    \sigma_t = \text{Sigmoid}(\text{FFN}([\hat{v}_0; \hat{r}_0])).
\end{equation}
The final navigation score for $V_i$ is:
\begin{equation}
    s_{i} = \sigma_t s^c_{i} + (1-\sigma_t)s^{f'}_{i}.
\end{equation}

\subsection{Training and Inference}
\label{sec:method_training}
\noindent\textbf{Pretraining.}
As shown in \cite{hao2020towards,chen2021hamt,pashevich2021episodic}, it is beneficial to pretrain transformer-based VLN models with auxiliary tasks as initialization.
Therefore, we first pretrain our model based on off-line expert demonstrations with behavior cloning and other common vision-and-language proxy tasks.
We use masked language modeling (MLM) \cite{devlin2019bert}, masked region classification (MRC) \cite{lu2019vilbert}, single-step action prediction (SAP) \cite{chen2021hamt} and object grounding (OG) \cite{lin2021scene} if object annotations are available.
The SAP and OG loss in behavior cloning given a demonstration path $\mathcal{P}^{*}$ is as follows:
\begin{gather}
    L_{\text{SAP}} = \sum\nolimits_{t=1}^{T} - \text{log}~p(a^{*}_t|\mathcal{W}, \mathcal{P}^{*}_{<t})\\
    L_{\text{OG}} = -\text{log}~p(o^{*}|\mathcal{W}, \mathcal{P}_T)
\end{gather}
where $a^{*}_t$ is the expert action of a partial demonstration path $\mathcal{P}^{*}_{<t}$, and $o^{*}$ is the groundtruth object at the last location $\mathcal{P}_T$.
More details are presented in the supplementary material.

\noindent\textbf{Policy learning via an interactive demonstrator.}
Behavior cloning suffers from distribution shifts between training and testing. 
Therefore, we propose to further train the policy with the supervision from a pseudo interactive demonstrator (PID) $\pi^{*}$ similar to the DAgger algorithm \cite{ross2011reduction}.
During training we have access to the environment graph $\mathcal{G}$, hence $\pi^*$ can utilize $\mathcal{G}$ to select the next target node, i.e., a navigable node with the overall shortest distance from the current node and to the final destination.
In each iteration, we use the current policy to sample a trajectory $\mathcal{P}$ and use $\pi^{*}$ to obtain pseudo supervision:
\begin{equation}
    L_{\text{PID}} = \sum\nolimits_{t=1}^{T} -\text{log}~p(a^{\pi^*}_t|\mathcal{W},\mathcal{P}_{<t})
\end{equation}
where $a^{\pi^*}_t$ is our pseudo target at step $t$.
We combine the original expert demonstrations with our pseudo demonstrations in policy learning with a balance factor $\lambda$:
\begin{equation}
\label{eqn:finetune_losses}
    L = \lambda L_{\text{SAP}} +  L_{\text{PID}} + L_{\text{OG}}.
\end{equation}

\noindent\textbf{Inference.}
At each time step during testing, we update the topological map as introduced in Sec.~\ref{sec:method_mapping} and then predict a global action as explained in Sec.~\ref{sec:method_global_planning}.
If it is a navigation action, the shortest route planning module employs the Floyd algorithm to obtain a shortest path from the current node to the predicted node given the map, otherwise the agent stops at the current location.
The agent is forced to stop if it exceeds the maximum action steps.
In such case, it will return to a node with maximum stop probability as its final prediction.
At the stopped location, the agent selects an object with maximum object prediction score.

\section{Experiments}

\subsection{Datasets} 
We focus our evaluation on goal-oriented VLN benchmarks REVERIE~\cite{qi2020reverie} and SOON~\cite{zhu2021soon}, which require fine-grained object grounding and advanced exploration capabilities to find a remote object. We also evaluate our model on the widely used VLN benchmark R2R~\cite{anderson2018vision}, which has step-by-step instructions and no object localization.
\smallskip\\
\noindent\textbf{REVERIE} contains high-level instructions mainly describing target locations and objects. Instructions contain 21 words on average. Given predefined object bounding boxes provided for each panorama, the agent should select the correct object bounding box at the end of the navigation path. The length of expert paths ranges from 4 to 7 steps.
\smallskip\\
\noindent\textbf{SOON} also provides instructions describing target rooms and objects. The average length of instructions is 47 words. 
SOON does not provide object boxes and requires the agent to predict object center locations in the panorama. Hence, we use an automatic object detector~\cite{anderson2018bottom} to obtain candidate object boxes.
The length of expert paths ranges from 2 to 21 steps with 9.5 steps on average.
\smallskip\\
\noindent\textbf{R2R} contains step-by-step navigation instructions. The average length of instructions is 32 words.
The average length of expert paths is 6 steps.
\smallskip\\
Examples from REVERIE and R2R are illustrated in Figure~\ref{fig:examples}. 
Further details are in the supplementary material.

\subsection{Evaluation Metrics}
\noindent\textbf{Navigation metrics.} 
We use standard metrics~\cite{anderson2018evaluation} to measure navigation performance, i.e., Trajectory Length (TL): average path length in meters; Navigation Error (NE): average distance in meters between agent’s final location and the target; Success Rate (SR): the ratio of paths with NE less than 3 meters; Oracle SR (OSR): SR given oracle stop policy; and SR penalized by Path Length (SPL). 

\noindent\textbf{Object grounding metrics.} To evaluate both the navigation and object grounding, we follow~\cite{qi2020reverie} and adopt Remote Grounding Success (RGS): the proportion of successfully executed instructions.
We also use RGS penalized by Path Length (RGSPL).
All the metrics are the higher the better except for TL and NE.

\subsection{Implementation Details}
\noindent\textbf{Features.}
For images, we adopt ViT-B/16 \cite{dosovitskiy2020image} pretrained on ImageNet to extract features.
For objects, we use the same ViT on the REVERIE dataset as it provides bounding boxes, while we use the BUTD object detector \cite{anderson2018bottom} on the SOON dataset.
The orientation feature \cite{ma2019self} contains $\sin(\cdot)$ and $\cos(\cdot)$ values for heading and elevation angles.

\noindent\textbf{Model architecture.}
We use 9, 2, 4 and 4 transformer layers in the text encoder, panorama encoder, coarse-scale cross-modal encoder and fine-scale cross-modal encoder, respectively.
Other hyper-parameters are set the same as in LXMERT \cite{tan2019lxmert}, \eg, the hidden layer size is 768.
We utilize the pretrained LXMERT for initialization.

\noindent\textbf{Training details.}
On the REVERIE dataset, we first pretrain \duet~with the batch size of 32 for 100k iterations using 2 Nvidia Tesla P100 GPUs. We automatically generate synthetic instructions to augment the dataset \cite{fried2018speaker}.
Then we use Eq.~(\ref{eqn:finetune_losses}) to fine-tune the policy with the batch size of 8 for 20k iterations on a single Tesla P100.
The best epoch is selected by SPL on val unseen split.
More details are provided in supplementary material.

\begin{table}
	\centering
	\small
	\tabcolsep=0.11cm
	\caption{Comparison of different scales and dual-scale fusion strategy on REVERIE val unseen split.}
	\label{tab:ablation_multiscale}
	\begin{tabular}{cc|ccccccc} \toprule
		 scale & fusion & OSR$\uparrow$ & SR$\uparrow$ & $\frac{\text{SR}}{\text{OSR}}\uparrow$ & SPL$\uparrow$ & RGS$\uparrow$ & RGSPL$\uparrow$ \\ \midrule
		fine & - & 30.96 & 28.86 & \textbf{93.22} & 23.57 & 20.39 & 16.64 \\
		coarse & - & 46.44 & 36.52 & 78.64 & 25.98 & - & - \\  \midrule
		\multirow{2}{*}{multi} & average & \textbf{51.86} & 45.81 & 88.33 & 31.94 & \textbf{32.49} & 22.78 \\
		& dynamic & 51.07 & \textbf{46.98} & 91.40 & \textbf{33.73} & 32.15 & \textbf{23.03} \\ \bottomrule
	\end{tabular}		
	\vspace{-1em}
\end{table}

\subsection{Ablation Study}
We ablated our approach on the REVERIE dataset. All results in this section are reported on the val unseen split. 

\noindent\textbf{1) Coarse-scale vs. fine-scale encoders.}
We first evaluate coarse-scale and fine-scale encoders separately for the REVERIE navigation task in the upper part of Table~\ref{tab:ablation_multiscale}.
As the coarse-scale encoder is not fed with object representations, it is unable to select target objects for the REVERIE task.
However, it outperforms the fine-scale version except for $\frac{\text{SR}}{\text{OSR}}$, for which the fine-scale encoder achieves much higher performance. 
This ratio estimates the performance of the stop action (the OSR is the success rate under oracle stop policy) and  indicates that fine-grained visual representations are essential to determine the target location specified in the instruction.
However, the fine-scale encoder obtains a low OSR score, suggesting it lacks exploration due to a limited action space.
The coarse-scale encoder instead benefits from the constructed map and is able to efficiently explore more areas with high OSR and SPL metrics.

\noindent\textbf{2) Dual-scale fusion strategy.}
As the fine- and coarse-scale encoders are complementary, we compare different approaches to fuse the two encoders in the bottom part of Table~\ref{tab:ablation_multiscale}.
Both fusion methods outperform the fine-scale and coarse-scale encoder by a large margin.
Our proposed dynamic fusion achieves more efficient exploration compared to the average fusion with 1.79\% improvement on SPL.

\noindent\textbf{3) Graph-aware self-attention.}
Table~\ref{tab:ablation_graphaware} ablates models with or without graph topology encoded in the transformer as in Eq.~(\ref{eqn:graphaware_attn}).
It shows that the awareness of the graph structures is more beneficial to improve the SPL score, which emphasizes navigating to the target with shorter distance.

\begin{table}
\centering
\small
\tabcolsep=0.15cm
\caption{Ablation of graph-aware self-attention (GASA) for graph encoding on REVERIE val unseen split.}
\label{tab:ablation_graphaware}
\begin{tabular}{cc|ccccc} \toprule
Fusion & GASA & OSR$\uparrow$ & SR$\uparrow$ & SPL$\uparrow$ & RGS$\uparrow$ & RGSPL$\uparrow$ \\ \midrule
\multirow{2}{*}{average} & $\times$ & 49.22 & 44.50 & 30.90 & 29.88 & 20.73 \\
 & \checkmark & \textbf{51.86} & 45.81 & 31.94 & 32.49 & 22.78 \\ \midrule
\multirow{2}{*}{dynamic} & $\times$ & 49.25 & 45.24 & 32.88 & 29.91 & 21.57 \\
 & \checkmark & 51.07 & \textbf{46.98} & \textbf{33.73} & \textbf{32.15} & \textbf{23.03} \\ \bottomrule
\end{tabular}
\end{table}

\noindent\textbf{4) Training losses.}
In Table~\ref{tab:ablation_training}, we compare different training losses for \duet.
The first row only uses $L_{\text{SAP}}$ in behavior cloning. As it is not trained for object grounding, we can ignore RGS and RGSPL metrics.
The second row adds the object supervision in training. It also improves navigation performance, which suggests that additional cross-modal supervisions such as association between words and objects can be beneficial to VLN tasks.
In the third row, we add common auxiliary proxy tasks MLM and MRC in training, which are more helpful for object grounding. As instructions in REVERIE mainly describe the final target, these two losses are more relevant to object grounding.
We further fine-tune the model with reinforcement learning (RL) \cite{hong2020recurrent,chen2021hamt} or our PID in the last two rows to address distribution shift issue in behavior cloning.
Both RL and PID achieve significant improvement and PID outperforms RL.

\begin{table}
\centering
\small
\tabcolsep=0.08cm
\caption{Ablation of training losses on REVERIE val unseen split.}
\label{tab:ablation_training}
\begin{tabular}{ccccc|ccccc} \toprule
\multicolumn{3}{c}{Pretrain} & \multicolumn{2}{c|}{Finetune} & \multirow{2}{*}{OSR$\uparrow$} & \multirow{2}{*}{SR$\uparrow$} & \multirow{2}{*}{SPL$\uparrow$} & \multirow{2}{*}{RGS$\uparrow$} & \multirow{2}{*}{RGSPL$\uparrow$} \\
SAP & OG & Aux & RL & PID &  &  &  &  &  \\ \midrule
\checkmark & $\times$ & $\times$ & $\times$ & $\times$ & 38.45 & 35.30 & 24.55 & - & - \\
\checkmark & \checkmark & $\times$ & $\times$ & $\times$ & 40.24 & 37.80 & 26.40 & 23.89 & 16.36 \\
\checkmark & \checkmark & \checkmark & $\times$ &  $\times$ & 37.63 & 36.81 & 27.19 & 25.05 & 18.40 \\
\checkmark & \checkmark & \checkmark & \checkmark & $\times$ & 47.51 & 42.35 & 32.97 & 29.91 & \textbf{23.53} \\
\checkmark & \checkmark & \checkmark & $\times$ & \checkmark & \textbf{51.07} & \textbf{46.98} & \textbf{33.73} & \textbf{32.15} & 23.03 \\ \bottomrule
\end{tabular}
\end{table}

\noindent\textbf{5) Data augmentation with synthetic instructions.}
We evaluate contributions of augmenting training data with synthetic instructions.
The upper block of Table~\ref{tab:ablation_augdata} presents results of pretraining with or without the augmented data. We can see that the synthetic data is beneficial in the pretraining stage and improves SPL and RGSPL by 1.63\% and 1.76\% respectively.
Based on the initialization of the model in row 2, we use PID to further improve the policy. The results are shown in the bottom block of Table~\ref{tab:ablation_augdata}.
The synthetic data however does not bring improvements to the performance.
We hypothesize that auxiliary proxy tasks in pretraining help to take advantage from the noisy synthetic data, but the policy learning still requires cleaner data.

\begin{table}
\centering
\small
\caption{Ablation of augmented speaker data in training on REVERIE val unseen split.}
\label{tab:ablation_augdata}
\begin{tabular}{cc|ccccc} \toprule
PID & Aug & OSR$\uparrow$ & SR$\uparrow$ & SPL$\uparrow$ & RGS$\uparrow$ & RGSPL$\uparrow$ \\ \midrule
\multirow{2}{*}{$\times$} & $\times$ & 37.29 & 34.56 & 25.56 & 23.00 & 16.64 \\
 & \checkmark & 37.63 & 36.81 & 27.19 & 25.05 & 18.40 \\ \midrule
\multirow{2}{*}{\checkmark} & $\times$ & 51.07 & \textbf{46.98} & \textbf{33.73} & \textbf{32.15} & \textbf{23.03}  \\
 & \checkmark & \textbf{52.09} & 46.58 & 32.72 & 31.75 & 22.18 \\ \bottomrule
\end{tabular}
\end{table}

\begin{table}
\centering
\small
\tabcolsep=0.13cm
\caption{Comparison with the state of the art on SOON dataset.}
\label{tab:soon_sota_cmpr}
\begin{tabular}{c|l|ccccc} \toprule
Split & Methods & TL & OSR$\uparrow$ & SR$\uparrow$ & SPL$\uparrow$ & RGSPL$\uparrow$ \\ \midrule
\multirow{2}{*}{\begin{tabular}[c]{@{}c@{}}Val \\ Unseen\end{tabular}} & GBE \cite{zhu2021soon} & 28.96 & 28.54 & 19.52 & 13.34 & 1.16 \\
 & \duet~(Ours) & 36.20 & \textbf{50.91} & \textbf{36.28} & \textbf{22.58} & \textbf{3.75} \\ \midrule
\multirow{2}{*}{\begin{tabular}[c]{@{}c@{}}Test \\ Unseen\end{tabular}} & GBE \cite{zhu2021soon} & 27.88 & 21.45 & 12.90 & 9.23 & 0.45 \\
& \duet~(Ours) & 41.83 & \textbf{43.00} & \textbf{33.44} & \textbf{21.42} & \textbf{4.17} \\ \bottomrule
\end{tabular}
\end{table}

\begin{table*}
\footnotesize
\centering
\tabcolsep=0.07cm
\caption{Comparison with the state-of-the-art methods on REVERIE dataset.\vspace{-.3cm}}
\label{tab:reverie_sota_cmpr}
\begin{tabular}{l|cccccc|cccccc|cccccc} \toprule
\multirow{3}{*}{Methods} & \multicolumn{6}{c|}{Val Seen} & \multicolumn{6}{c|}{Val Unseen} & \multicolumn{6}{c}{Test Unseen} \\
\multicolumn{1}{c|}{} & \multicolumn{4}{c}{Navigation} & \multicolumn{2}{c|}{Grounding} & \multicolumn{4}{c}{Navigation} & \multicolumn{2}{c|}{Grounding} & \multicolumn{4}{c}{Navigation} & \multicolumn{2}{c}{Grounding}  \\ 
 & TL & OSR$\uparrow$ & SR$\uparrow$ & SPL$\uparrow$ & RGS$\uparrow$ & RGSPL$\uparrow$ & TL & OSR$\uparrow$ & SR$\uparrow$ & SPL$\uparrow$ & RGS$\uparrow$ & RGSPL$\uparrow$ & TL & OSR$\uparrow$ & SR$\uparrow$ & SPL$\uparrow$ & RGS$\uparrow$ & RGSPL$\uparrow$ \\ \midrule
Human & - & - & - & - & - & - & - & - & - & - & - & - & 21.18 & 86.83 & 81.51 & 53.66 & 77.84 & 51.44 \\ \midrule
Seq2Seq \cite{anderson2018vision} & 12.88 & 35.70 & 29.59 & 24.01 & 18.97 & 14.96 & 11.07 & 8.07 & 4.20 & 2.84 & 2.16 & 1.63 & 10.89 & 6.88 & 3.99 & 3.09 & 2.00 & 1.58 \\
RCM \cite{wang2019reinforced} & 10.70 & 29.44 & 23.33 & 21.82 & 16.23 & 15.36 & 11.98 & 14.23 & 9.29 & 6.97 & 4.89 & 3.89 & 10.60 & 11.68 & 7.84 & 6.67 & 3.67 & 3.14 \\
SMNA \cite{ma2019self} & 7.54 & 43.29 & 41.25 & 39.61 & 30.07 & 28.98 & 9.07 & 11.28 & 8.15 & 6.44 & 4.54 & 3.61 & 9.23 & 8.39 & 5.80 & 4.53 & 3.10 & 2.39 \\
FAST\scriptsize{-MATTN} \cite{qi2020reverie} & 16.35 & 55.17 & 50.53 & 45.50 & 31.97 & 29.66 & 45.28 & 28.20 & 14.40 & 7.19 & 7.84 & 4.67 & 39.05 & 30.63 & 19.88 & 11.61 & 11.28 & 6.08 \\
SIA \cite{lin2021scene} & 13.61 & 65.85 & 61.91 & 57.08 & 45.96 & 42.65 & 41.53 & 44.67 & 31.53 & 16.28 & 22.41 & 11.56 & 48.61 & 44.56 & 30.80 & 14.85 & 19.02 & 9.20 \\
RecBERT \cite{hong2020recurrent} & 13.44 & 53.90 & 51.79 & 47.96 & 38.23 & 35.61 & 16.78 & 35.02 & 30.67 & 24.90 & 18.77 & 15.27 & 15.86 & 32.91 & 29.61 & 23.99 & 16.50 & 13.51 \\ 
Airbert \cite{guhur2021airbert} & 15.16 & 48.98 & 47.01 & 42.34 & 32.75 & 30.01 & 18.71 & 34.51 & 27.89 & 21.88 & 18.23 & 14.18 & 17.91 & 34.20 & 30.28 & 23.61  & 16.83 & 13.28 \\
HAMT \cite{chen2021hamt} & 12.79 & 47.65 & 43.29 & 40.19 & 27.20 & 25.18 & 14.08 & 36.84 & 32.95 & 30.20 & 18.92 & 17.28 & 13.62 & 33.41 & 30.40 & 26.67 & 14.88 & 13.08 \\ \midrule
\duet~(Ours) & 13.86 & \textbf{73.86} & \textbf{71.75} & \textbf{63.94} & \textbf{57.41} & \textbf{51.14} & 22.11 & \textbf{51.07} & \textbf{46.98} & \textbf{33.73} & \textbf{32.15} & \textbf{23.03} & 21.30 & \textbf{56.91} & \textbf{52.51} & \textbf{36.06} & \textbf{31.88} & \textbf{22.06} \\ \bottomrule
\end{tabular}
\end{table*}

\subsection{Comparison with State of the Art}

\noindent\textbf{REVERIE.}
Table~\ref{tab:reverie_sota_cmpr} compares our final model with state-of-the-art models on the REVERIE dataset.
Our model significantly beats the state of the arts on all evaluation metrics on the three splits. For example, on the val unseen split, our model outperforms the previous best model HAMT \cite{chen2021hamt} by 14.03\% on SR, 3.53\% on SPL and 5.75\% on RGSPL. Our model also generalizes better on the test unseen split, where we improve over HAMT by {\color{red}22.11\%} on SR, {\color{red}9.39\%} on SPL and {\color{red}8.98\%} on RGSPL.
This clearly demonstrates the effectiveness of our dual-scale action planning model with topological maps. Note that none of the previous methods has employed a map for navigation on this dataset.

\noindent\textbf{SOON.}
Table~\ref{tab:soon_sota_cmpr} presents the results on the SOON dataset.
Our model also achieves significant better performance than the previous graph-based approach GBE \cite{zhu2021soon}, with {\color{red}20.54\%} gains on SR and {\color{red}12.19\%} on SPL on test unseen split. 
The results, however, are much lower than those on REVERIE. This is because SOON contains fewer and more challenging training data (see supplementary material for analysis).

\begin{figure}
    \centering
    \begin{overpic}[width=1\linewidth]{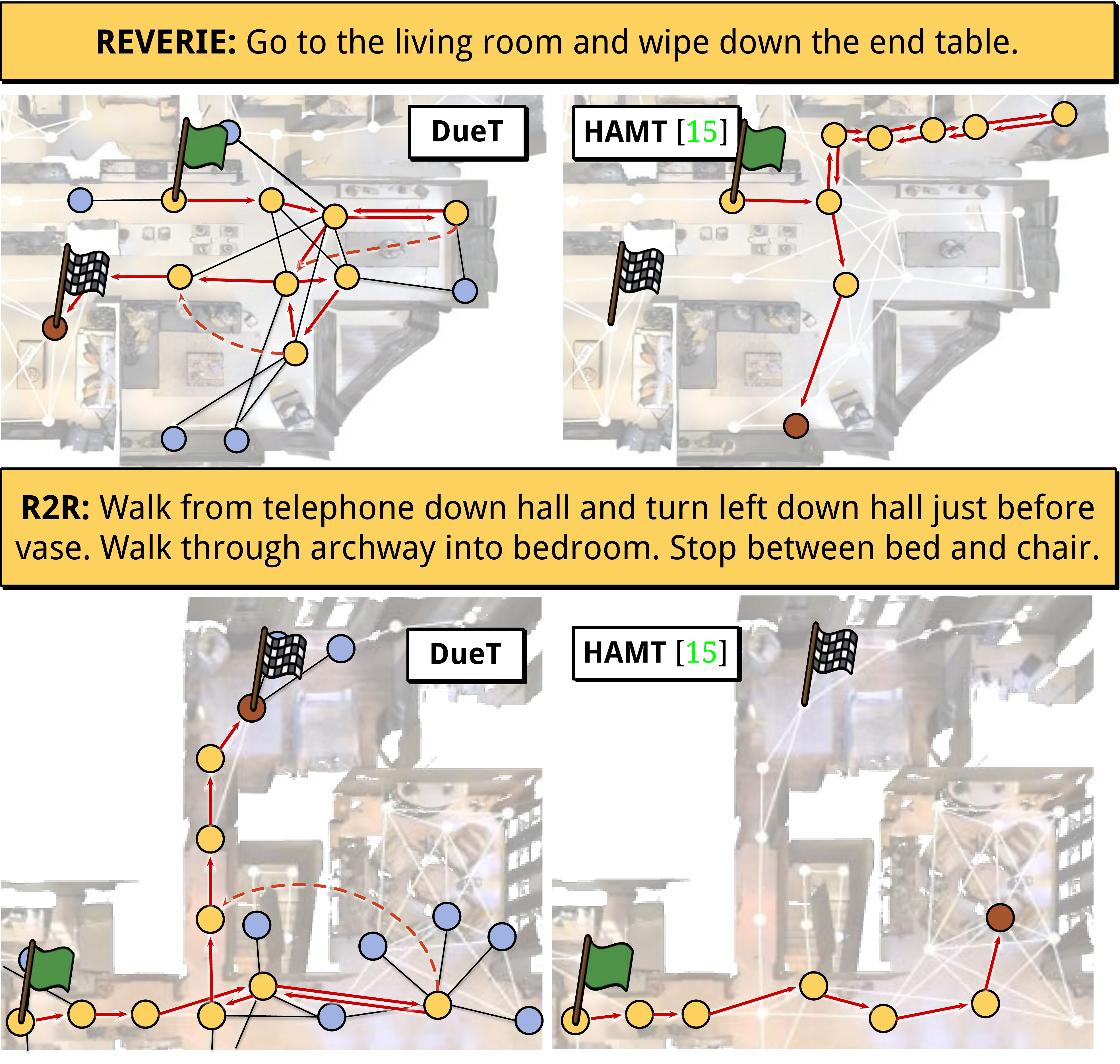}
    \end{overpic}
    \caption{Predicted trajectories of \duet~and the state-of-the-art HAMT \cite{chen2021hamt}. The green and checkered flags denote start and target locations respectively. The dashed lines denote global actions. \duet~is able to make more efficient explorations and correct its previous decisions, while HAMT is limited by its local actions.}
    \label{fig:examples}
    \vspace{-1.5em}
\end{figure}

\begin{table}
	\centering
	\footnotesize
	\tabcolsep=0.07cm
	\caption{Comparison with the state of the art on R2R dataset. Methods are grouped according to the used memories: `Rec' for recurrent state, `Seq' for sequence and `Map' for topological map.}
	\label{tab:r2r_sota_cmpr}
	\begin{tabular}{l|l|cccccccc} \toprule
		\multirow{2}{*}{Mem} & \multirow{2}{*}{Methods} &  \multicolumn{4}{c}{Val Unseen} & \multicolumn{4}{c}{Test Unseen} \\
		& & TL$\downarrow$ & NE$\downarrow$ & SR$\uparrow$ & SPL$\uparrow$ & TL$\downarrow$ & NE$\downarrow$ & SR$\uparrow$ & SPL$\uparrow$  \\ \midrule
		\multirow{8}{*}{Rec} & Seq2Seq \cite{anderson2018vision} & 8.39 & 7.81 & 22 & - & 8.13 & 7.85 & 20 & 18 \\
		& SF \cite{fried2018speaker} & - & 6.62 & 35 & - & 14.82 & 6.62 & 35 & 28 \\
		& PRESS \cite{li2019robust} & 10.36 & 5.28 & 49 & 45 & 10.77 & 5.49 & 49 & 45 \\
		& EnvDrop \cite{tan2019learning} & 10.70 & 5.22 & 52 & 48 & 11.66 & 5.23 & 51 & 47 \\
		& AuxRN \cite{zhu2020vision} & - & 5.28 & 55 & 50 & - & 5.15 & 55 & 51 \\
		& {\scriptsize PREVALENT} \cite{hao2020towards} & 10.19 & 4.71 & 58 & 53 & 10.51 & 5.30 & 54 & 51 \\
		& RelGraph \cite{hong2020language} & 9.99 & 4.73 & 57 & 53 & 10.29 & 4.75 & 55 & 52 \\
		& RecBERT \cite{hong2020recurrent} & 12.01 & 3.93 & 63 & 57 & 12.35 & 4.09 & 63 & 57 \\  \midrule
        \multirow{2}{*}{Seq} & HAMT \cite{chen2021hamt} & 11.87 & 3.65 & 65 & 59 & 12.65 & 4.11 & 63 & 58 \\
		& HAMT{\scriptsize-e2e} \cite{chen2021hamt} & 11.46 & \textbf{2.29} & 66 & \textbf{61} & 12.27 & 3.93 & 65 & \textbf{60} \\ \midrule
        \multirow{5}{*}{Map} & EGP \cite{deng2020evolving} & - & 4.83 & 56 & 44 & - & 5.34 & 53 & 42 \\ 
        & GBE \cite{zhu2021soon} & - & 5.20 & 54 & 43 & - & 5.18 & 53 & 43 \\
        & SSM \cite{wang2021structured} & 20.7 & 4.32 & 62 & 45 & 20.4 & 4.57 & 61 & 46 \\ \cmidrule{2-10}
        & \duet-coarse & 12.96 & 3.67 & 68 & 59 & 13.08 & 3.93 & 67 & 58 \\
		& \duet~(Ours) & 13.94 & 3.31 & \textbf{72} & 60 & 14.73 & \textbf{3.65} & \textbf{69} & 59 \\ \bottomrule
	\end{tabular}
\end{table}

\noindent\textbf{R2R.}
As shown in Table~\ref{tab:r2r_sota_cmpr}, \duet~beats state-of-the-art approaches on success rate (SR) by 6\% and 4\% on val unseen and test unseen split respectively. However, it achieves comparable performances on SPL. This can be explained by the fact that for map-based approaches backtracking is encouraged which makes the trajectory length longer. 
We further compare a coarse-scale \duet~for fair comparison with previous graph-based approaches~\cite{zhu2021soon,deng2020evolving,wang2021structured} which do not use a fine-scale encoder. 
Even without using the fine-scale representation, \duet~still outperform them by a margin, showing the effectiveness of our graph transformer.
It also demonstrates \duet~is able to backtrack more efficiently. 
Figure~\ref{fig:examples} visualizes some qualitative examples.

\section{Conclusion}
We propose \duet~(dual-scale graph transformer) for vision-and-language navigation (VLN) based on online constructed topological maps.
It uses graph transformers to reason over a coarse-scale map representation for long-term action planning and a fine-scale local representation for fine-grained language grounding.
The two scales are dynamically combined in the navigation policy.
\duet~achieves state-of-the-art performance on VLN benchmarks REVERIE, SOON and R2R.
However, our approach is not always successful as demonstrated by the gap between seen and unseen environments, and is restricted to discrete environments. Future work will address these points. 
Applications of our work should take security and privacy risks into account.

\noindent\textbf{Acknowledgement.}
This work was granted access to the HPC resources of IDRIS under the allocation 101002 made by GENCI. 
This work is funded in part by the French government under management of Agence Nationale de la Recherche as part of the ``Investissements d'avenir'' program, reference ANR19-P3IA-0001 (PRAIRIE 3IA Institute) and by Louis Vuitton ENS Chair on Artificial Intelligence.

\appendix

\section*{Appendix}

Section~\ref{sec:model_details} provides additional details for the model. 
The experimental setup is described in Section~\ref{sec:expr_details}, including datasets, metrics and implementation details.
Section~\ref{sec:supp_ablation} presents more ablation studies of our \duet~model.
Section~\ref{sec:examples} shows more qualitative examples.

\section{Model Details}
\label{sec:model_details}

\subsection{Pretraining Objectives}
As introduced in Sec~\ref{sec:method_training}, %
we employ two auxiliary proxy tasks in pretraining in addition to behavior cloning tasks SAP (single-step action prediction) and OG (object grounding).
In the following, we describe the two auxiliary tasks: masked language modeling (MLM) and masked region classification (MRC). 
The inputs for the two tasks are pairs of instruction $\mathcal{W}$ and demonstration path $\mathcal{P}$.

\smallskip

\noindent\textbf{Masked Language Modeling (MLM)} task aims to learn grounded language representations and cross-modal alignment by predicting masked words given contextual words and demonstration path.
We randomly replace tokens in $\mathcal{W}$ by a special token \verb|[mask]| with the probability of 15\% \cite{devlin2019bert}.
Both the coarse-scale encoder and fine-scale encoder can generate contextual word embeddings for masked words as introduced in Sec~\ref{sec:coarse_scale_encoding} and \ref{sec:fine_scale_encoding} respectively. %
The coarse-scale encoder utilizes visual information from an encoded graph at the final step as contexts, while the fine-scale encoder utilizes the last panoramic observation as visual contexts.
We average output embeddings of the two encoders for masked words, and employ a two-layer fully-connected network to predict word distributions $p(w_i|\mathcal{W}_{\backslash i}, \mathcal{P})$ where $\mathcal{W}_{\backslash i}$ is the masked instruction and $w_i$ is the label of masked word. The objective of the task is minimizing the negative log-likelihood of original words: $L_{\text{MLM}} = - \mathrm{log}\ p (w_i|\mathcal{W}_{\backslash i}, \mathcal{P})$.

\smallskip

\noindent\textbf{Masked Region Classification (MRC)} aims to predict semantic labels of masked image regions in an observation given an instruction and neighboring regions.
As instructions in goal-oriented VLN tasks mainly describe the last observation in the demonstration path, we only apply the MRC task on the fine-scale encoder.
We randomly zero out view images and objects in the last observation of $\mathcal{P}$ with the probability of 15\%. 
The target semantic labels for view images are class probability predicted by an image classification model \cite{dosovitskiy2020image} pretrained on ImageNet, while the labels for objects are class probability predicted by an object detector \cite{anderson2018bottom} pretrained on VisualGenome.
We use a two-layer fully-connected network to predict semantic labels for each masked visual token, and minimize the KL divergence between the predicted and target probability distribution.

\begin{table*}
	\centering
	\small
	\caption{Dataset statistics. \#house, \#instr denote the number of houses and instructions respectively.}
	\label{tab:dataset_stats}
	\begin{tabular}{cccccccccc} \toprule
		\multirow{2}{*}{VLN Task} & \multirow{2}{*}{Dataset} & \multicolumn{2}{c}{Train} & \multicolumn{2}{c}{Val Seen} & \multicolumn{2}{c}{Val Unseen} & \multicolumn{2}{c}{Test Unseen} \\
		& & \#house & \#instr & \#house & \#instr & \#house & \#instr & \#house & \#instr \\ \midrule
		\multirow{2}{*}{Object-oriented} & REVERIE \cite{qi2020reverie} & 60 & 10,466 & 46 & 1,423 & 10 & 3,521 & 16 & 6,292 \\
		& SOON \cite{zhu2021soon} & 34 & 2,780 & 2 & 113 &  5 & 339 & 14 & 1,411 \\ \midrule
		\multirow{2}{*}{Fine-grained} & R2R \cite{anderson2018vision} & 61 & 14,039 & 56 & 1,021 & 11 & 2,349 & 18 & 4,173 \\
		& R4R \cite{jain2019stay} & 59 & 233,532 & 40 & 1,035 & 11 & 45,234 & - & - \\ \bottomrule
	\end{tabular}
\end{table*}

\subsection{Speaker Model for Data Augmentation}
We train a speaker model to synthesize instructions based on visual observations for REVERIE dataset.
As REVERIE provides annotated object classes and Matterport3D also contains annotated room classes, we utilize these semantic labels to alleviate the gap between vision and language.
Our speaker model consists of a panorama encoder and a sentence decoder.
The panorama encoder is fed with image features of the panorama, semantic labels of target object and target room as well as the level of the room.
We project all the input features into the same dimension, and utilize a transformer with self-attention to capture relations of each token.
The sentence decoder then sequentially generates words conditioning on the encoded tokens.
We use LSTM as the decoder and follow the architecture in show-attend-tell image captioning model \cite{xu2015show}.

Please note that we only employ data in REVERIE training split to learn the speaker model.
We initialize the word embeddings in encoder and decoder with pretrained GloVe embeddings \cite{pennington2014glove} and train the speaker model for 50 epochs.
We employ the trained speaker model to synthesize instructions for every annotated object in the REVERIE training split, leading to 19,636 instructions in total.
We extend the size of the training set from 10,466 instruction-path pairs to 30,102 pairs.

\section{Experimental Setups}
\label{sec:expr_details}

\subsection{Dataset}

We primarily focus our evaluation on goal-oriented VLN benchmarks REVERIE~\cite{qi2020reverie} and SOON~\cite{zhu2021soon}. To localize target objects in these benchmarks, the agent requires fine-grained object grounding and advanced exploration capabilities. We also test our model on less demanding VLN benchmarks R2R~\cite{anderson2018vision} with step-by-step instructions and no object localization.
All the benchmarks build upon the Matterport3D~\cite{chang2017matterport3d} environment and contain 90 photo-realistic houses.
Each house is defined by a set of navigable locations. Each location is represented by the corresponding panorama image, GPS coordinates and a set of possible actions. We adopt the standard split of houses into {\em training}, {\em val seen}, {\em val unseen}, and {\em test} subsets. Houses in the {\em val seen} split are the same as in {\em training}, while houses in {\em val unseen} and {\em test} splits are different from {\em training}.

Table~\ref{tab:dataset_stats} presents statistics of the three datasets. To be noted, we follow the released challenge split on SOON dataset instead of the split in the original paper \cite{zhu2021soon}\footnote{As shown in \url{https://github.com/ZhuFengdaaa/SOON/issues/1}, Zhu \etal \cite{zhu2021soon} do not release the split in their original paper. Therefore, performance comparisons on SOON dataset are based on their challenge report \url{https://scenario-oriented-object-navigation.github.io/}.}.

\subsection{Data Processing for SOON Dataset}
The SOON dataset does not provide annotated object bounding boxes per panorama. It only annotates the location of target object bounding boxes for each instruction, including the orientation of object's center point as well as orientation of top left, top right, bottom left, and bottom right corners.
The object grounding setting in SOON dataset is to predict the orientation of object's center point.
However, we observe that though the annotated objects' center points are of good quality, their annotations of the four corners are quite noisy\footnote{As shown in \url{https://github.com/ZhuFengdaaa/SOON/issues/2}, about 50\% polygons constructed by the annotated four corners do not contain the objects' center point.}.
Therefore, we propose to clean the object bounding boxes in training and also provide more automatically detected objects as fine-grained visual contexts to represent each panorama.

Specifically, we employ the BUTD detector \cite{anderson2018bottom} pretrained on VisualGenome to detect objects per panorama, which covers 1600 object and scene classes. We filter some unimportant classes for SOON dataset such as `background', `floor',  `ceiling', `wall', `roof' and so on.
We then select one of the detected objects as our pseudo target according to the semantic similarity of object classes and the Euclidean distances of the objects' center points compared to annotated target object.
In this way, we convert the object grounding setting in SOON datset similar to the setting in REVERIE dataset, whose goal is to select one object from all candidate objects.
In inference, we utilize the orientation of the selected object as our object grounding prediction.

\subsection{Evaluation Metrics}
Due to the different settings for object grounding in REVERIE and SOON datasets, definitions of success in the two datasets are different.
In REVERIE dataset, the success is defined as arriving at a location where the target object is visible and selecting the target object among all annotated candidate objects in the panorama of the location.
In SOON dataset, an agent succeeded in carrying out an instruction if it arrives 3 meters near to one of the target locations and the predicted orientation of target object's center point is inside of the annotated polygon of the object in the location.

\subsection{Training Details}
\noindent\textbf{REVERIE}:
In pretraining, we combine the original dataset with augmented data synthesized by our speaker model. We pretrain \duet~with the batch size of 32 for 100k iterations using 2 Nvidia Tesla P100 GPUs. 
Then we use Eq.~(12) presented in the main paper to fine-tune the policy with the batch size of 8 for 20k iterations on a single Tesla P100.
The best epoch is selected by SPL on val unseen split.

\noindent\textbf{SOON}:
As the size of SOON dataset is much smaller than REVERIE dataset and  the instructions are much more complicated, we do not synthesize instructions for SOON dataset.
We pretrain model using the original instructions and our automatically cleaned object bounding boxes for 40k iterations with batch size of 32.
We fine-tune the model for 40k iterations with batch size of 2 on a single Tesla P100 and select the best model by SPL on val unseen split.

\noindent\textbf{R2R}:
Following previous works \cite{hao2020towards,hong2020recurrent,chen2021hamt}, we adopt augmented R2R data \cite{hao2020towards} in pretraining. We pretrain the model for 200k interations with batch size of 64.
We fine-tune the model for 20k iterations with batch size of 8.

\begin{table}
\centering
\caption{Ablation of balance factor $\lambda$ in the fine-tuning loss.}
\label{tab:supp_abl_lambda_loss}
\begin{tabular}{c|ccccc} \toprule
\multirow{2}{*}{} & \multicolumn{3}{c}{Navigation} & \multicolumn{2}{c}{Object Grounding} \\
 & OSR & SR & SPL & RGS & RGSPL \\ \midrule
0 & \textbf{53.00} & \textbf{48.22} & 33.00 & 32.12 & 22.04 \\
0.2 & 51.07 & 46.98 & \textbf{33.73} & 32.15 & \textbf{23.03} \\
0.5 & 52.06 & 46.98 & 32.38 & \textbf{32.43} & 22.72 \\
1 & 50.33 & 45.64 & 32.54 & 30.19 & 21.50 \\ \bottomrule
\end{tabular}
\end{table}

\section{Additional Ablations}
\label{sec:supp_ablation}
\subsection{Balance factor $\lambda$ in fine-tuning objective}
Table~\ref{tab:supp_abl_lambda_loss} presents the performance of using different $\lambda$ in the fine-tuning objective in Eq.~(12) of the main paper. The larger $\lambda$, the more important of the behavior cloning. 
We can see that over-emphasizing behavior cloning is harmful to the exploration ability. The model with $\lambda=1$ achieves the worst OSR and SR.
Removing behavior cloning ($\lambda=0$) achieves good navigation performance such as in OSR, SR and SPL, but it is less competitive in object grounding.
We think this is because the agent fails to navigate to target locations in its sampled trajectories, and is unable to train the object grounding module. However, the agent is guaranteed to arrive at target locations in behavior cloning.

\begin{figure}
    \centering
    \begin{overpic}[width=1\linewidth]{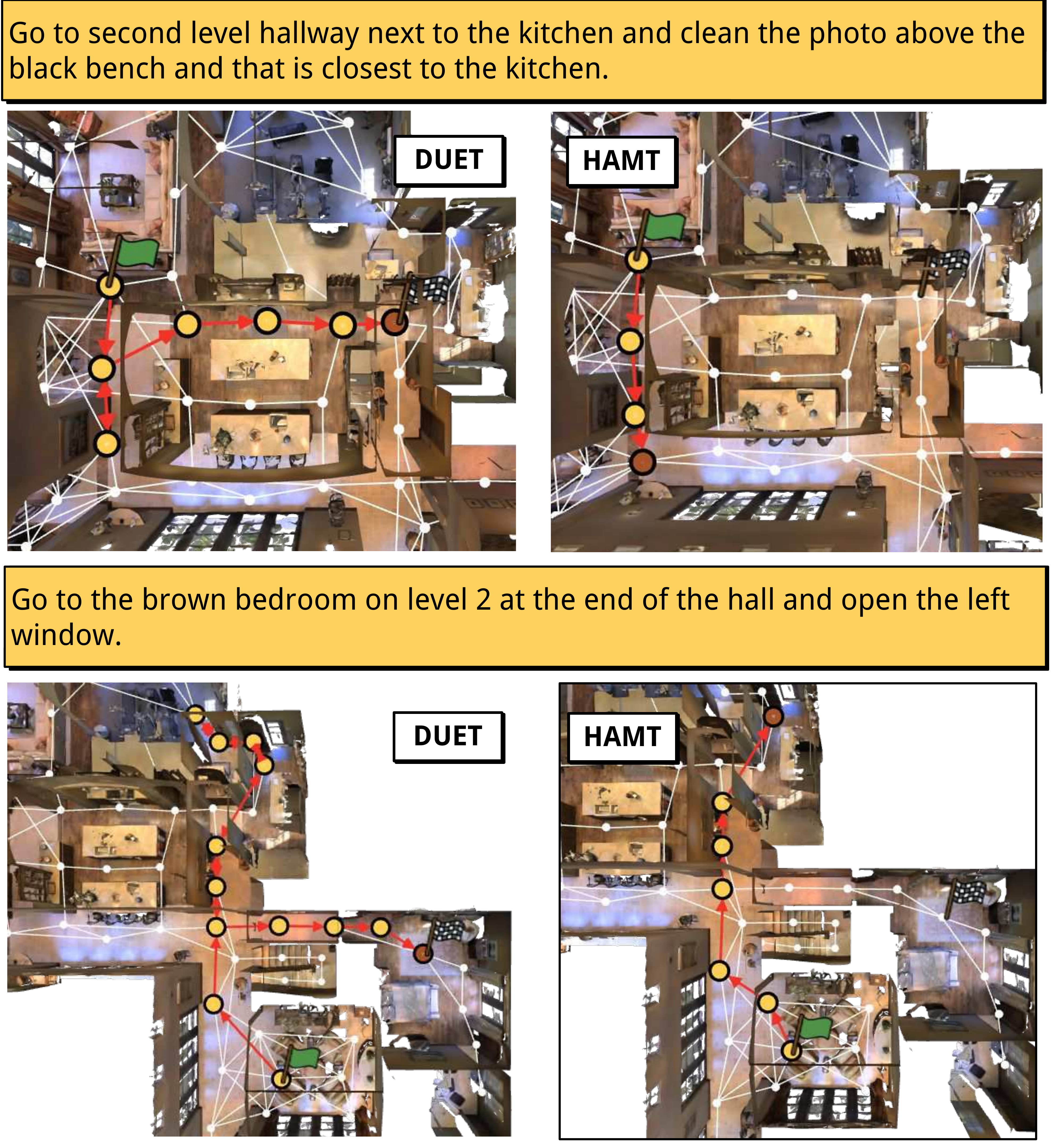}
    \end{overpic}
    \caption{Predicted trajectories of \duet~and the state-of-the-art HAMT \cite{chen2021hamt} on REVERIE val unseen split. The green and checkered flags denote start and target locations respectively.}
    \label{fig:supp_reverie_examples}
\end{figure}

\begin{figure}
    \centering
    \begin{overpic}[width=1\linewidth]{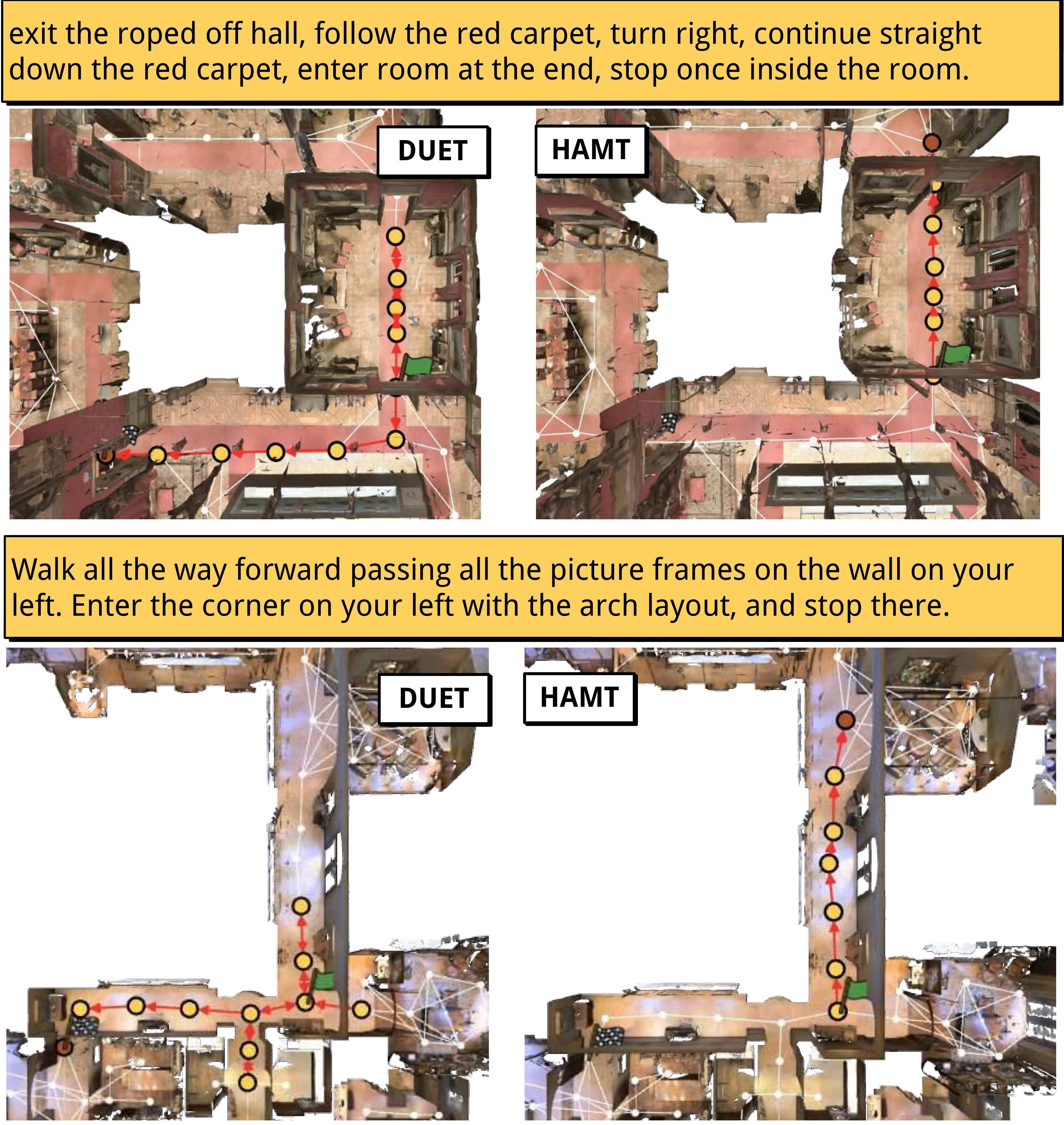}
    \end{overpic}
    \caption{Predicted trajectories of \duet~and the state-of-the-art HAMT \cite{chen2021hamt} on R2R val unseen split. The green and checkered flags denote start and target locations respectively.}
    \label{fig:supp_r2r_examples}
\end{figure}

\subsection{Backtrack ratio in inference}
The backtrack action indicates that the agent does not select a neighboring node from the local action space but jumps to a previously partially observed node through the global action space.
We compute the backtrack ratio for \duet. On the REVERIE val seen split, \duet~only backtracks in 13.7\% of the predicted trajectories; while on the REVERIE val unseen split, \duet~backtracks in 48.6\% of its predicted trajectories.
As the agent has the capacity to memorize house structures in seen environments, it can directly find the target location without much exploration in seen environments. However, when the agent is deployed in unseen environments, it has to explore more to find the target location specified by high-level instructions.
When step-by-step instructions are given such as in R2R dataset, we observe the backtrack ratio significantly decreases to 23.2\% on val unseen split, which matches our expectation.

\subsection{Fusion weights of coarse and fine scales}
We observe that the agent typically puts more weights on the fine-scale module in the beginning and at the end of the navigation, and on the coarse-scale module in the middle.
Quantitatively, the average weight of the coarse-scale module is 0.36 in the beginning, 0.45 in the middle, and 0.42 at the end.
The agent may not need to backtrack at early steps, so it relies more on the local fine-scale module. Then, the agent needs to explore so the global coarse-scale module gets more attention. When deciding where to stop, the agent should identify the target object and the fine-scale module is emphasized again.

\subsection{Failure analysis}
We perform an additional quantitative evaluation on the REVERIE dataset. For navigation, we measure whether an agent stops at the target room type (\eg a bathroom) or at the correct location. We obtain the following results: (a) incorrect room type: 29.82\%;
(b) correct room type + incorrect location: 23.20\%; (c) correct location: 46.98\%.
This shows that fine-grained scene understanding remains challenging.
With respect to object grounding, once an agent reaches the correct location, the object can be correctly localized 68.43\% of the time.

\begin{figure*}
    \centering
    \begin{overpic}[width=0.9\linewidth]{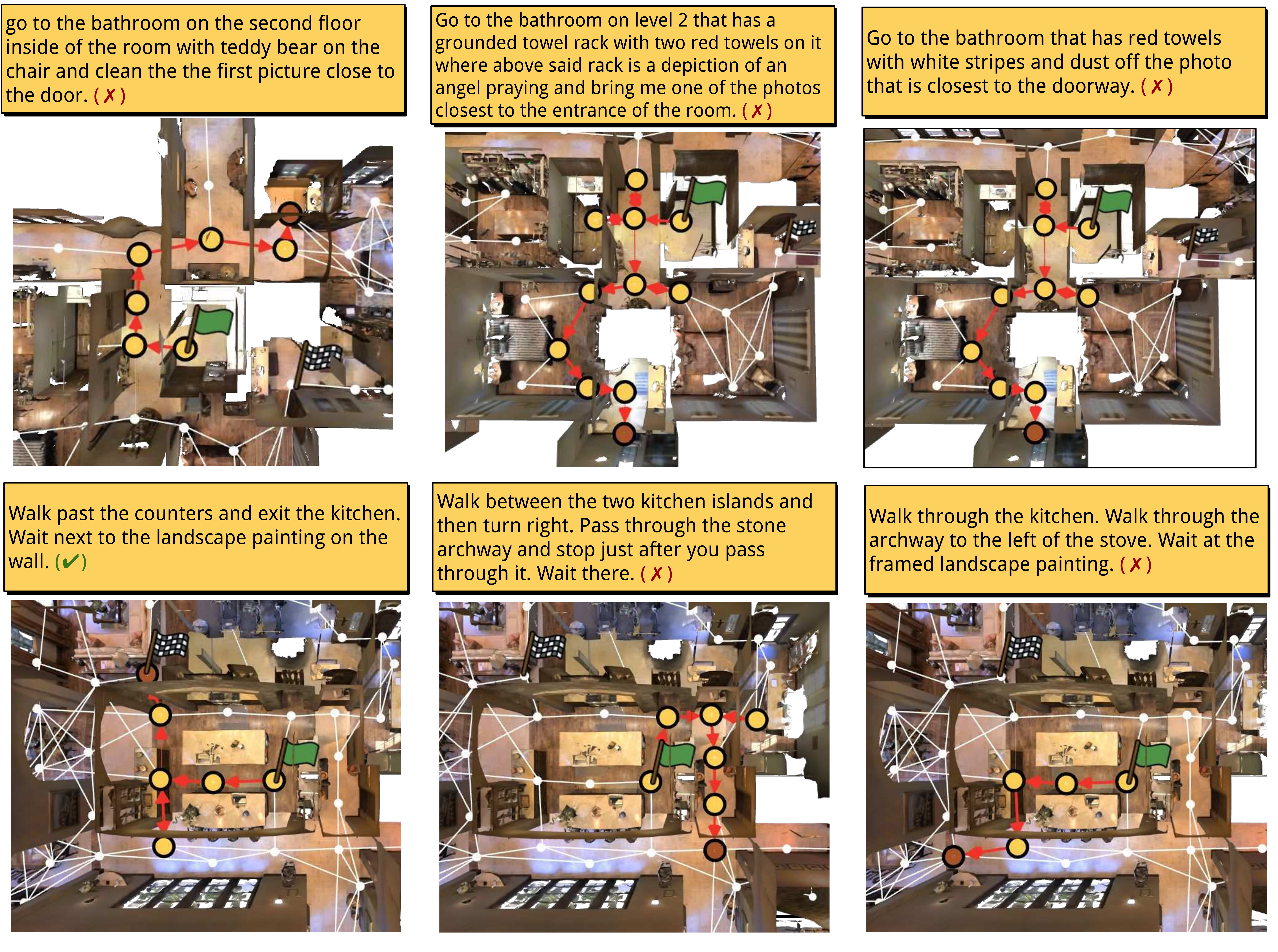}
    \end{overpic}
    \caption{Predicted trajectories of \duet~ on REVERIE val unseen split (top) and R2R val unseen split (bottom). The green and checkered flags denote start and target locations respectively.}
    \label{fig:supp_failure_cases}
\end{figure*}

\section{Qualitative Examples}
\label{sec:examples}

Figure~\ref{fig:supp_reverie_examples} visualizes some examples of our \duet~ and the state-of-the-art HAMT \cite{chen2021hamt} model on REVERIE dataset.
In both the cases, the agents explore an incorrect direction in the first attempt. However, \duet~is able to efficiently explore another direction towards the goal.
Figure~\ref{fig:supp_r2r_examples} shows some examples on R2R dataset. Though step-by-step instructions are provided, the instruction can still be ambiguous. For example, both directions of the start point in the top example of Figure~\ref{fig:supp_r2r_examples} can ``exit the rope off hall''.
\duet~is also better at correcting its previous decisions when it finds that the followup instructions do not match with the visual observations.

We further provide some failure cases in REVERIE and R2R datasets in Figure~\ref{fig:supp_failure_cases}.
In the top example of Figure~\ref{fig:supp_failure_cases}, there are several bathrooms in the house and our \duet~model arrives at one of bathroom. However, the arrived bathroom does not contain the fine-grained objects specified in the instruction. It suggests that our model still needs to improve the fine-grained object grounding capability.
The bottom example presents three different instructions for the same trajectory on R2R dataset. The agent succeeds in following the first instruction, but fails for the other two instructions. We observe that the predictions are not very robust across different language instructions.

\end{document}